\documentclass[a4paper]{article}
\usepackage[preprint]{neurips_2024}

\usepackage{graphicx}
\usepackage{times}
\usepackage{latexsym}
\usepackage[hyphens]{url} 
\usepackage{hyperref}
\usepackage{caption} 
\usepackage{array} 
\usepackage{booktabs}
\usepackage{float}
\usepackage{comment}
\usepackage{amsmath,amsthm,amsfonts,amssymb,bm,stmaryrd}
\usepackage{makecell} 
\usepackage{multirow, multicol}
\usepackage{colortbl}
\usepackage{xcolor}
\usepackage{wrapfig} 
\usepackage{pdflscape}
\usepackage[most]{tcolorbox}
\usepackage{makecell}
\usepackage{subcaption}

\title{Grouped Differential Attention}
\author{Motif Technologies}

\begin{document}

\maketitle

\begin{abstract}
The self-attention mechanism, while foundational to modern Transformer architectures, suffers from a critical inefficiency: it frequently allocates substantial attention to redundant or noisy context. Differential Attention addressed this by using subtractive attention maps for signal and noise, but its required balanced head allocation imposes rigid constraints on representational flexibility and scalability.

To overcome this, we propose Grouped Differential Attention (GDA), a novel approach that introduces unbalanced head allocation between signal-preserving and noise-control groups. GDA significantly enhances signal focus by strategically assigning more heads to signal extraction and fewer to noise-control, stabilizing the latter through controlled repetition (akin to GQA). This design achieves stronger signal fidelity with minimal computational overhead. We further extend this principle to group-differentiated growth, a scalable strategy that selectively replicates only the signal-focused heads, thereby ensuring efficient capacity expansion.

Through large-scale pretraining and continual training experiments, we demonstrate that moderate imbalance ratios in GDA yield substantial improvements in generalization and stability compared to symmetric baselines. Our results collectively establish that ratio-aware head allocation and selective expansion offer an effective and practical path toward designing scalable, computation-efficient Transformer architectures.
\end{abstract}

\section{Introduction}

% https://arxiv.org/pdf/2502.07864 여기 있는 그림이나 https://arxiv.org/pdf/2305.13245 여기 Figure 2 참조해서 개념 그림 하나 있으면 좋을 듯 함.

Transformer-based models, first introduced in the seminal work~\cite{vaswani2017attention}, have fundamentally reshaped the landscape of modern AI, becoming the foundational backbone of large-scale models across diverse fields like natural language processing, computer vision, and multimodal learning. This pervasive success is largely attributable to the architecture's exceptional capacity for sequence modeling and representation learning.

Despite its power, a persistent and inherent limitation plagues the core self-attention mechanism: its frequent failure to reliably focus on key information. This often results in the generation of redundant or noisy interactions among tokens, which can dilute the quality of the learned representations. To effectively counteract this challenge, recent research has explored various architectural remedies. For example, large-scale models like GPT-OSS~\cite{agarwal2025gpt} have incorporated attention sink mechanisms to stabilize the attention process and actively suppress activations deemed irrelevant. A more structural approach, known as Differential Attention \cite{ye2024differential}, was introduced as a promising extension to mitigate this noise. This design elegantly operates by constructing two complementary attention maps: one dedicated to computing meaningful signals and the other focused on capturing the incidental noises. The core innovation lies in combining these two maps through subtraction, effectively performing a``noise removal'' operation on the captured signals. This disentanglement of signal from noise has been shown to significantly enhance the robustness and stability of model training.

However, the original design of Differential Attention harbors a critical inefficiency. It mandates a symmetric allocation of attention heads across the two groups -- the signal-computing heads and the noise-control heads. This enforced, balanced split implicitly assumes that signal extraction and noise reduction are tasks of equal importance and require identical resource allocation. Consequently, this symmetric approach often leads to an excessive allocation of computational resources to the noise group. This redundancy, in turn, reduces the capacity available for the primary function of capturing meaningful semantic dependencies, thereby limiting the model's overall expressivity. As a direct result, Differential Transformers have seen limited widespread adoption in the large-scale training environment, apart from a few isolated successes such as in the Motif 2.6B model \cite{lim2025motif}. This observation highlights a clear architectural necessity: to develop a transformer design that retains the structural advantages of noise suppression while simultaneously avoiding these resource inefficiencies, reducing redundant noise computations, and achieving better scalability without sacrificing representational expressivity.

In this paper, we propose a novel approach to address these inefficiencies: Grouped Differential Attention (GDA). GDA is an architectural generalization that extends Differential Attention by supporting an unbalanced, asymmetric allocation of heads between the signal-preserving and noise-control groups. Instead of being constrained to an even division, our design strategically allocates a larger share of attention heads to the signal-preserving group. Concurrently, the noise-control group operates with a reduced capacity, which is structurally stabilized through controlled repetition. This stabilization technique is conceptually akin to the approach successfully employed in Grouped Query Attention (GQA) \cite{ainslie2023gqa}. This deliberate architectural bias ensures the model leans toward stronger, more comprehensive signal extraction while still maintaining the crucial functionality of structural noise suppression. Our central hypothesis is that this imbalance, when implemented moderately, enables a more effective and efficient use of the model's representational capacity within any given fixed computational budget.

Furthermore, we extend this principle of asymmetry from a static head allocation to a more dynamic scaling mechanism via \emph{group-differentiated growth}. This presents a powerful alternative to the conventional method of uniform hypercloning, where all heads are expanded proportionally. Our asymmetric approach selectively replicates only the signal-preserving heads, leaving the noise-control heads unreplicated or minimally expanded. This asymmetric expansion is key: it unlocks significant additional representational power by focusing resource growth where it matters most -- the signal extraction -- thereby avoiding the introduction of unnecessary computational redundancy. This offers a more flexible and streamlined pathway for continual growth when transitioning from smaller to significantly larger models.

We validate the efficacy of these concepts through extensive, large-scale pretraining experiments across a diverse set of challenging benchmarks, including those focused on reasoning, commonsense, and general knowledge. The empirical results unequivocally highlight the decisive role played by the allocation ratio between the signal-preserving and noise-control heads. Specifically, we find that moderate imbalance ratios, e.g., 3:1 or 4:1, strike the optimal balance between powerful signal extraction and necessary noise suppression. These moderate ratios consistently yield stronger generalization performance compared to the original, symmetrically designed baselines. These compelling findings lead us to a clear conclusion: an effective and efficient differential transformer design fundamentally hinges on maintaining an appropriate, moderate balance between signal capacity and noise control, thereby ensuring the achievement of both high efficiency and robust scalability.

\section{Methodology}

\subsection{Preliminary}

\begin{figure}[t]
    \centering
    \includegraphics[width=0.85\linewidth]{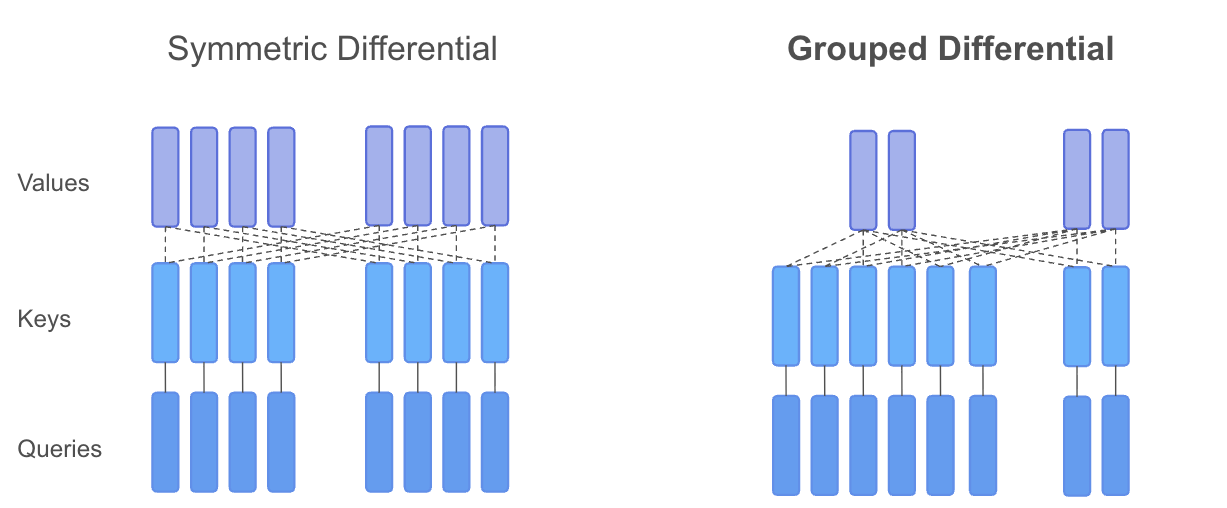}
    \caption{
        Overview of the \textit{Grouped Differential Attention (GDA)}.
        Unlike the symmetric Differential Transformer, GDA allocates heads unevenly across signal and noise groups,
        sharing a smaller set of noise-control heads among multiple query groups.
    }
    \label{fig:gdt_overview}
\end{figure}
Differential Attention \cite{ye2024differential} extends the standard self-attention mechanism by enhancing attention to relevant context while suppressing irrelevant signals. In their formulation, attention noise is defined as the phenomenon in which excessive attention scores are allocated to irrelevant context. To mitigate this, Differential Attention constructs two complementary attention maps and subtracts one from the other, thereby neutralizing noisy patterns in the scores. The resulting formulation in the multi-head setting is presented below.

Let the model dimension be $d_{\text{model}}$, and suppose that we use $H$ attention heads. 
Then, each head has projection matrices 
$W_Q^{i}, W_K^{i}, W_V^{i} \in \mathbb{R}^{d_{\text{model}} \times 2d_{\text{head}}}$ 
with $d_{\text{head}} = d_{\text{model}} / H$, 
and the input sequence $X \in \mathbb{R}^{N \times d_{\text{model}}}$ 
is projected into query, key, and value representations. 
for each $i$-th head as follows:

\begin{equation}
\begin{aligned}
\relax [Q^{i}_1 ; Q^{i}_2] &= X W_Q^{i}, 
\quad [K^{i}_1 ; K^{i}_2] = X W_K^{i}, 
\quad V^{i} = X W_V^{i} .
\end{aligned}
\end{equation}

Intuitively, the heads are split evenly into two groups, i.e., a $1{:}1$ partition, so that each group contains \(h = \frac{H}{2}\) heads. For each head, two query-key pairs $(Q^{i}_1, K^{i}_1)$ and $(Q^{i}_2, K^{i}_2)$ are generated by linear projections of the input $X$. These pairs yield two parallel attention maps. Note that the value projection $V$ is shared across both branches, allowing the two attention maps to focus on distinct weighting patterns over the same content space rather than learning separate feature representations. Unlike standard scaled dot-product attention, differential attention combines the two maps by subtraction, so that one branch emphasizes signal while the other cancels noise. 
The balance between the two attention maps is governed by a learnable scalar  
\[
\lambda = \exp(\boldsymbol{\lambda_{q_1}} \cdot \boldsymbol{\lambda_{k_1}}) 
- \exp(\boldsymbol{\lambda_{q_2}} \cdot \boldsymbol{\lambda_{k_2}}) 
+ \lambda_{\text{init}},
\]
where $\boldsymbol{\lambda_{q_i}}$ and $\boldsymbol{\lambda_{k_i}}$ are learnable parameter vectors and $\lambda_{\text{init}}$ is a fixed scalar.

Formally, the output of the $i$-th head is defined as
\begin{equation}
\begin{aligned}
\mathrm{head}_i 
&= \Bigg(
    \underbrace{\mathrm{softmax}\!\Big(\tfrac{Q^{i}_1 (K^{i}_1)^{\top}}{\sqrt{d_h}}\Big)}_{\text{signal map}}
    \;-\;
    \lambda \cdot 
    \underbrace{\mathrm{softmax}\!\Big(\tfrac{Q^{i}_2 (K^{i}_2)^{\top}}{\sqrt{d_h}}\Big)}_{\text{noise map}}
\Bigg) V^{i}, \\[4pt]
\overline{\mathrm{head}_i} 
&= (1 - \lambda_{\text{init}}) \cdot \operatorname{LN}(\mathrm{head}_i).
\end{aligned}
\end{equation}

Here, $\operatorname{LN}(\cdot)$ denotes Root Mean Square normalization~\cite{zhang2019root} applied to each head. Finally, we combine the outputs from all heads using concatenation followed by a linear projection, yielding the multi-head differential attention output:

\begin{equation}
\begin{aligned}
\operatorname{MultiHeadAttn}(X) &= \operatorname{Concat}(\overline{\mathrm{head}}_1, \ldots, \overline{\mathrm{head}}_h) W_O
\end{aligned}
\end{equation}

where the output projection is $W_O \in \mathbb{R}^{Hd_{\text{head}} \times d_{\text{model}}}$,
and $\operatorname{Concat}(\cdot)$ concatenates the heads together along the channel dimension. 

\subsection{Grouped Differential Attention}

We extend the Differential Transformer’s paradigm of noise cancellation with a novel unbalanced group allocation scheme. Rather than evenly splitting attention heads across two parallel softmax streams, we assign much capacity to the “signal-preserving” group. Specifically, fewer attention heads are assigned to the noise-cancelling group, whose reduced capacity is compensated through controlled repetition , which is essentially equivalent to the head-sharing mechanism of Grouped Query Attention (GQA) \cite{ainslie2023gqa}. The majority of heads are thus devoted to capturing signals, while the residual noise contributes to increased feature diversity. Intuitively, this suggests that an appropriate balance between the two groups is essential for effective representation learning.

\paragraph{Grouping of Heads} 
The input queries, keys, and values are partitioned into two distinct groups: 
\((Q_1, Q_2)\), \((K_1, K_2)\). 
The first group \((Q_1, K_1)\) is dedicated to \emph{signal-focused computation}, 
while the second group \((Q_2, K_2)\) functions as a \emph{noise-control buffer}, 
as in the standard Differential Transformer. 
Unlike the symmetric allocation in the original formulation, 
we explicitly adopt an unbalanced group ratio, for example
\begin{equation}
    Q_1 : Q_2 \;=\; G : 1, 
    \qquad G > 1,
\end{equation}
which allocates proportionally greater representational capacity to the signal group. This skew ensures that high-fidelity attention patterns dominate the aggregated output, while the noise group, operating at reduced dimensionality, leverages controlled redundancy, for examples, repeating or reshaping projections, to stabilize learning without introducing significant overhead.  

Formally, let the number of attention heads be \(H\), evenly divided into \(G+1\) groups, with each group containing $h=$\(H/(G+1)\) heads. For head index \(i \in \{0, \ldots, H-h\}\), the group index is defined as
\begin{equation}
    g_i = \left\lfloor \frac{i}{h} \right\rfloor .
\end{equation}
Two heads \(i\) and \(k\) belong to the same group if and only if
\begin{equation}
    \left\lfloor \frac{i}{h} \right\rfloor 
    = \left\lfloor \frac{k}{h} \right\rfloor .
\end{equation}

The output of head \(i\) is then given by
\begin{equation}
    \mathrm{head}_i
    = \Bigg( 
        \mathrm{softmax}\!\Big( 
            \tfrac{Q^{i}_1 (K^{i}_1)^{\top}}{\sqrt{d_h}} 
        \Big)
        - \lambda \cdot 
        \mathrm{softmax}\!\Big( 
            \tfrac{Q^{g_i}_2 (K^{g_i}_2)^{\top}}{\sqrt{d_h}} 
        \Big) 
    \Bigg) V^{g_i} ,
\end{equation}
where \(Q^{g_i}_2,\), \(K^{g_i}_2\) and \(V^{g_i}\) denote the matrices associated with the group \(g_i\). Each value projection \(W_V^{g}\) maps the shared input representation into a group-level content space, \(W_V^{g} \in \mathbb{R}^{d_{\text{model}} \times 2h d_{\text{head}}}\). Unlike the query and key projections, which are split into separate branches for signal and noise computation, the value projection remains unified within each group to ensure that both attention maps operate over a common content basis. This aggregation follows the same formulation as in the standard multi-head attention, differing only in the output projection dimension $W_O \in \mathbb{R}^{2(H-h)d_{\text{head}} \times d_{\text{model}}}$
\begin{equation}
\begin{aligned}
\operatorname{MultiHeadAttn}(X) &= \operatorname{Concat}(\overline{\mathrm{head}}_1, \ldots, \overline{\mathrm{head}}_{(H-h)}) W_O
\end{aligned}
\end{equation}

\begin{figure}[t]
    \centering
    
    % (a) Signal vs Noise Heads
    \begin{subfigure}[t]{0.45\linewidth}
        \centering
        \includegraphics[height=4cm]{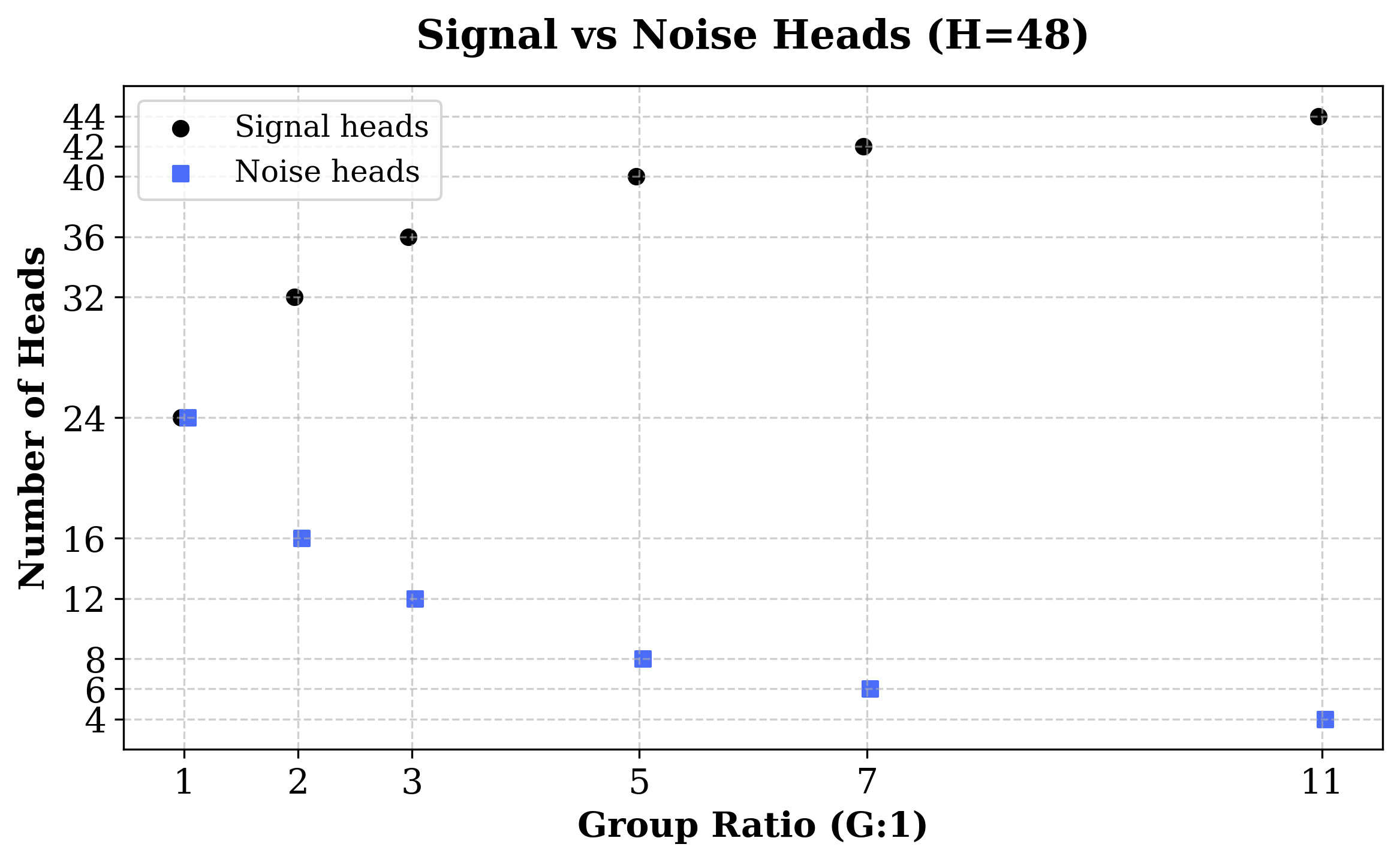}
        \caption{Head allocation across groups for $H=48$.}
        \label{fig:signal-vs-noise}
    \end{subfigure}
    \hfill
    % (b) Performance vs Ratio
    \begin{subfigure}[t]{0.45\linewidth}
        \centering
        \includegraphics[height=4cm]{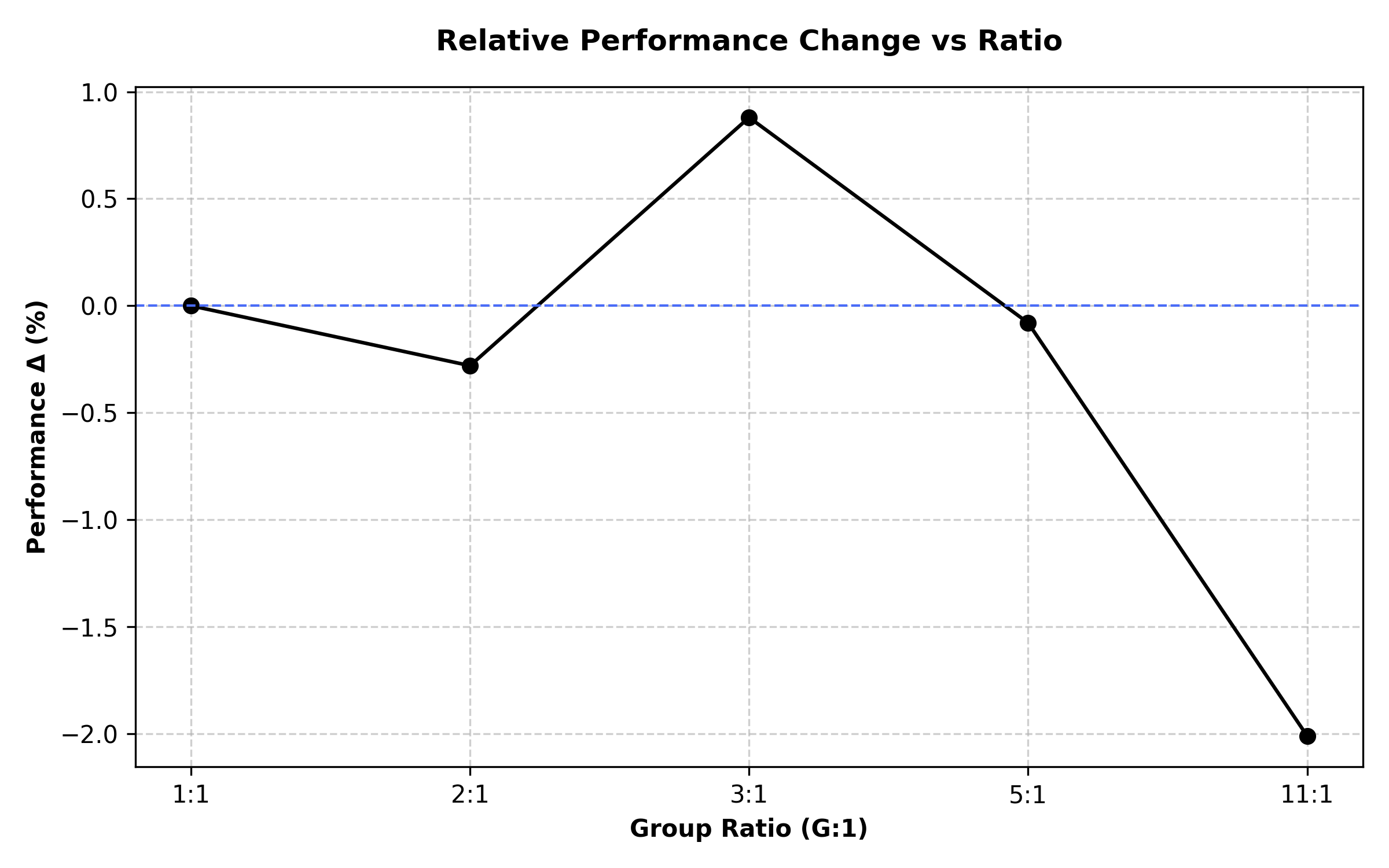}
        \caption{Relative performance change across group ratios.}
        \label{fig:performance-vs-ratio}
    \end{subfigure}
    
    \caption{Comparison of group-differentiated allocation and performance. 
    (a) Signal and noise head allocation under different $G{:}1$ ratios for $H=48$, 
    where $(G+1)$ divides $H$. 
    (b) Relative performance change (percentage $\Delta$) compared to the $1{:}1$ baseline.}
    \label{fig:allocation-performance}
\end{figure}

\paragraph{Imbalanced Allocation}
In contrast to the original Differential Transformer, where heads are divided symmetrically across two groups, a $1{:}1$ split yields $H/2$ effective output heads, our formulation introduces an \emph{imbalanced allocation}. Under a general $G{:}1$ ratio, the signal-preserving group is assigned 
$H - \tfrac{H}{G+1}$ heads, while the noise-control group receives only $\tfrac{H}{G+1}$ heads. 
This design increases the number of effective output heads by biasing capacity toward the signal-preserving group. To maintain computational efficiency within the same FLOPs budget, the number of value heads is reduced proportionally to the smaller noise-control group. Thus, the signal group gains greater representational strength without additional computational overhead, as illustrated in Figure~\ref{fig:signal-vs-noise}.

\section{Experiments}

We evaluate the proposed Transformer architecture for large-scale language models from two complementary perspectives. First, we compare it against the original Differential Transformer baselines under varying head grouping ratios, with the goal of identifying the most effective allocation of head capacity given a fixed FLOPs budget and training token count.

Second, we investigate \emph{progressive continual training}, in which models are incrementally scaled from smaller to larger sizes. This setting allows us to assess whether the proposed architecture offers advantages over uniform scaling 
when adapting to staged growth.

\subsection{Language Modeling Pretraining Evaluation}

\paragraph{Models and Dataset}

We conducted pretraining experiments on models with approximately 0.9B non-embedding parameters, trained under several configurations of the proposed grouped differential architecture with Group Query Attention. Each model utilizes a hidden dimension of 1536, 24 layers, 48 total attention heads, 12 key–value heads, and a head dimension of 32. We set the RoPE base to $\theta=10{,}000$ and used a maximum sequence length of 4096 tokens.
\begin{table}[H]
    \centering
    \resizebox{\linewidth}{!}{
    \begin{tabular}{l r r r r r}
        \toprule
        \textbf{Ratio} &
        \textbf{Signal Heads} &
        \textbf{Noise Heads} &
        \textbf{Query Heads (total)} &
        \textbf{Key–Value Heads} &
        \textbf{Layers} \\
        \midrule
        1:1 (baseline) & 24 & 24 & 48 & 12 & 24  \\
        2:1            & 32 & 16 & 48 & 12 & 24  \\
        3:1            & 36 & 12 & 48 & 12 & 24  \\
        5:1            & 40 &  8 & 48 & 12 & 24  \\
        11:1           & 44 &  4 & 48 & 12 & 24  \\
        \bottomrule
    \end{tabular}
    }
    \caption{Model configurations under varying imbalance ratios. Each model is trained with the same depth, hidden size, and activation, differing only in the allocation of signal versus noise heads.}
    \label{tab:model_config_ratios}
\end{table}
The key distinction among experimental variants lies in the allocation of \emph{noise heads}. We explored a range of head group ratios, from \textbf{1:1 (baseline)} to \textbf{11:1}, corresponding to 24 down to 4 noise heads while maintaining a fixed total of 48 attention heads.

Training was performed on a corpus emphasizing broad domain coverage to promote generalization. Our corpus consists of large-scale web data from Common Crawl \cite{commoncrawl}, mathematical content from FineMath \cite{finemath2025hf}, and additional open-source reasoning datasets. Optimization was performed using AdamW~\cite{loshchilov2017decoupled} with a learning rate of \(5 \times 10^{-4}\), \((\beta_1, \beta_2) = (0.9, 0.95)\), weight decay of 0.1, and a 5\% warmup schedule. Pretraining used a global batch size of 4M tokens. We applied a Warmup–Stable–Decay (WSD) learning-rate scheduler \cite{hu2024minicpmunveilingpotentialsmall}, with the decay phase limited to 10\% of the total training steps.

\paragraph{Results}  

We evaluated the models in a few-shot setting using benchmarks spanning mathematics, question answering, and commonsense reasoning, including PIQA \cite{bisk2020piqa}, ARC (Challenge/Easy)~\cite{clark2018arc}, HellaSwag~\cite{zellers2019hellaswag}, MMLU~\cite{hendrycks2021mmlu}, MMLU-Pro \cite{wang2024mmlupro}, and GSM8K \cite{cobbe2021gsm8k}.

As shown in Table~\ref{tab:fromscratch_results}, moderate imbalance ratios produced slight improvements on several tasks while maintaining stable performance on MMLU and related evaluations. In contrast, GSM8K performance declined, which we attribute to the relatively limited proportion of mathematical content in the training corpus. At extreme imbalance ratios, performance consistently degraded across benchmarks, indicating inefficient parameter utilization. Overall, these findings suggest that mild imbalance can be beneficial, whereas excessive asymmetry reduces generalization and weakens capacity allocation. The relative performance trends across different head-group ratios are further illustrated in Figure~\ref{fig:allocation-performance}b.

\begin{table}[H]
    \centering
    \resizebox{\linewidth}{!}{
    \begin{tabular}{l r r r r r r r r }
        \toprule
        \textbf{Ratio} &
        \textbf{PIQA} &
        \textbf{ARC Challenge} &
        \textbf{ARC Easy} &
        \textbf{Hellaswag} &
        \textbf{MMLU} &
        \textbf{MMLU Pro} &
        \textbf{GSM8K} &
        \textbf{Avg Gain (\%)} \\
        \midrule
        1:1 (baseline) & 71.22 & 30.72 & 58.67 & 50.14 & 31.28 & 10.95 & 4.70 & 0 \\
        2:1            & 71.60 & 29.86 & 59.30 & 49.94 & 31.43 & 11.64 & 3.18 & -0.28 \\
        3:1            & 71.44 & 31.31 & 59.93 & 49.95 & 31.49 & 11.81 & 4.02 & +0.88 \\
        5:1            & 71.60 & 31.31 & 59.18 & 49.41 & 30.97 & 11.43 & 3.71 & -0.08 \\
        11:1           & 70.89 & 29.18 & 59.64 & 48.36 & 31.05 & 11.18 & 2.20 & -2.01 \\
        \bottomrule
    \end{tabular}
    }
    \caption{Benchmark results under varying imbalance ratios. Performance is reported on commonsense reasoning (PIQA, ARC, Hellaswag), knowledge-intensive evaluation (MMLU, MMLU Pro), and mathematical reasoning (GSM8K). Average gain is normalized to the baseline.}
    \label{tab:fromscratch_results}
\end{table}
\subsection{Progressive Continual Training}

\paragraph{Configuration}
Our second line of evaluation investigates \emph{progressive continual training}, in which smaller models are incrementally scaled into larger ones through multiple stages. As a baseline, we apply the \emph{HyperCloning} method \cite{samragh2024scaling} to the standard Differential Transformer, where the parameters of linear layers are replicated by an integer factor without slicing. This approach preserves the forward computation, namely, the output logits, while constraining replication to integer multiples to maintain structural consistency during expansion. In contrast, we investigate \emph{group-differentiated growth}, which allows head groups to expand at different rates by replicating signal-preserving heads while leaving noise-control heads not replicated. This flexible strategy consistently yields stronger performance and improved training stability, providing a more efficient and robust pathway for scaling that balances computational cost with stable learning dynamics.

We began with a Differential Transformer comprising approximately 400M non-embedding parameters. To investigate scalability, we doubled the hidden dimension, leading to a fourfold increase in effective capacity (about 1.6B parameters). The training configuration followed the same setup as the pretraining from scratch described in Section~3.1.

\begin{table}[t]
    \centering
    \resizebox{\linewidth}{!}{
    \begin{tabular}{lcccccccc}
        \toprule
        \textbf{Configuration} & \textbf{Hidden Dim} & \textbf{Layers} & \textbf{Total Heads} & \textbf{Noise Heads} & \textbf{KV Heads} & \textbf{Head Dim} & \textbf{Expand} & \textbf{Rep. Factor} \\
        \midrule
        Baseline (Small, Ratio 1:1)              & 1024 & 24 & 16 & 8  & 16 & 64 & False    & 1 \\
        Scaled Baseline (Ratio 1:1, Hypercloning) & 2048 & 24 & 32 & 16 & 16 & 64 & True  & 2 \\
        Group-Diff (Ratio 3:1, Hypercloning)                   & 2048 & 24 & 32 & 8  & 16 & 64 & True  & 2 \\
        Group-Diff (Ratio 4:1, Hypercloning)                   & 2048 & 24 & 40 & 8  & 16 & 64 & True  & 2 \\
        \bottomrule
    \end{tabular}
    }
    \caption{Model configurations for baseline and group-differentiated growth experiments.}
    \label{tab:model_configs}
\end{table}

To evaluate allocation strategies, we compared several head-group ratios, including the balanced baseline (1:1) and skewed allocations (3:1 and 4:1). Ratios beyond this range were not explored, as earlier results indicated that excessively unbalanced configurations tend to degrade performance (cf. Section~3.1). Based on these findings, we hypothesized that the optimal range lies between 3:1 and 5:1, and accordingly conducted experiments with 3:1 and 4:1 ratios under appropriately scaled head settings. These configurations enabled a systematic study of how head-group ratios influence performance under both direct scaling and hyperclone-based expansion.

Training was carried out in two stages: an initial 160B tokens on the small-scale model, followed by an additional 80B tokens using \emph{progressive continual training}. This staged setup allowed us to assess the effectiveness of gradual scaling—building upon a strong base model—versus training a larger model entirely from scratch. Most training configurations were kept consistent with Section~3.1, unless otherwise specified.

\paragraph{Results}
Among the evaluated configurations, the 4:1 ratio achieved the strongest performance, surpassing both the balanced baseline and the moderately skewed 3:1 setting. This outcome suggests that, under hypercloning-based replication, disproportionately enlarging the signal-dominant group enhances representational capacity and yields measurable gains on benchmarks, as shown in Table~\ref{tab:imbalance_results}. These findings further support the utility of \emph{group-differentiated growth}, where head groups expand at distinct rates.

By moving beyond simple replication, this approach enables more effective structural allocation, thereby improving both efficiency and training stability. The advantages of progressive/growth-based scaling align with insights from progressive network methods and elastic architectures in other domains \cite{gong2019efficient,expanse2022progressive}. Altogether, these results indicate that group-differentiated growth offers a more stable and effective pathway for scale-up in large language models.

\begin{table}[H]
    \centering
    \resizebox{\linewidth}{!}{
    \begin{tabular}{l r r r r r r r r }
        \toprule
        \textbf{Ratio} &
        \textbf{PIQA} &
        \textbf{ARC Challenge} &
        \textbf{ARC Easy} &
        \textbf{HellaSwag} &
        \textbf{MMLU} &
        \textbf{MMLU Pro} &
        \textbf{GSM8K} &
        \textbf{Avg Gain (\%)} \\
        \midrule
        1:1 & 73.5  & 32.83 & 63.89 & 56.93 & 32.74 & 11.66 & 7.42  & 0.0 \\
        3:1 & 73.12 & 33.36 & 63.55 & 57.46 & 33.42 & 11.62 & 10.46 & +1.44 \\
        4:1 & 73.72 & 34.47 & 64.35 & 58.26 & 33.41 & 10.95 & 10.92 & +2.54 \\
        \bottomrule
    \end{tabular}
    }
    \caption{Evaluation results under progressive continual training with different head-group ratios. Gains are reported relative to the balanced baseline (1:1).}
    \label{tab:imbalance_results}
\end{table}

\section{Conclusion}

Our experimental results demonstrate that the relative allocation of attention heads is a decisive factor in how a model's representational capacity is distributed. We have shown that a moderate imbalance in the ratio between signal-preserving and noise-control heads can significantly strengthen generalization and model performance, particularly under fixed computational budgets. Conversely, an excessive degree of asymmetry consistently degrades both efficiency and robustness.

Beyond static ratio design, we introduced and validated a strategy for group-differentiated growth. This approach selectively expands the signal-focused heads while keeping the noise-control heads unreplicated. This technique proves to be a more effective and resource-aware alternative to simple, uniform head replication. By strategically unlocking capacity and focusing growth on the most impactful components, this method not only helps preserve previously learned representations during scaling but also leads to smoother optimization and stronger benchmark performance.

Taken together, our findings establish ratio-aware allocation and progressive, group-differentiated growth as highly practical and effective principles for scaling robust and efficient Transformer models.

\paragraph{Limitations}

Our study was conducted with certain constraints that warrant discussion. Specifically, our training budget was limited, falling well short of the trillion-token scale typically employed for state-of-the-art pretraining efforts. Furthermore, our evaluation was confined to a narrow set of reasoning and knowledge benchmarks. Future work should broaden the scope to include more challenging scenarios, such as long-context understanding, and involve a more detailed architectural analysis, including statistical examination of attention score distributions.

\bibliographystyle{plain}
\bibliography{reference}

\clearpage
\section{Appendix}

\subsection{Contribution}
All authors are sorted alphabetically by last name.

\textbf{Technical and management leadership}: Junghwan Lim

\textbf{Core contributors}: Sungmin Lee, Dongseok Kim

\textbf{Contributors}: Wai Ting Cheung, Beomgyu Kim, Taehwan Kim, Haesol Lee, Junhyeok Lee, Dongpin Oh, Eunhwan Park

\end{document}